# Knowledge Graph Completion Method Combined With Adaptive Enhanced Semantic Information

Weidong Ji [1,] Zengxiang Yin [1*], Guohui Zhou [1], Yuqi Yue[1], Xinru Zhang[1], Chenghong Sun[1]

[1] College of Computer Science and Information Engineering, Harbin Normal University, Harbin 150025, China;
kingjwd@126.com(W,J);zhouguohui@hrbnu.edu.cn(G,Z);yueyuqi980717@163.com (Y,Y)
zhangxinru990604@163.com (X,Z); sunchenghong1998@163.com(C,S);
* Correspondence: yinzengxiang@163.com(Z,Y)

**Conflicts of Interest:** The authors declare no conflict of interest.

**Acknowledgments:** The authors would like to thank all anonymous reviewers for their comments.

**Abstract:** Translation models tend to ignore the rich semantic information in triads in the process of knowledge graph complementation. To remedy this shortcoming, this paper constructs a knowledge graph complementation method that incorporates adaptively enhanced semantic information. The hidden semantic information inherent in the triad is obtained by fine-tuning the BERT model, and the attention feature embedding method is used to calculate the semantic attention scores between relations and entities in positive and negative triads and incorporate them into the structural information to form a soft constraint rule for semantic information. The rule is added to the original translation model to realize the adaptive enhancement of semantic information. In addition, the method takes into account the effect of high-dimensional vectors on the effect, and uses the BERT-whitening method to reduce the dimensionality and generate a more efficient semantic vector representation. After experimental comparison, the proposed method performs better on both FB15K and WIN18 datasets, with a numerical improvement of about 2.6% compared with the original translation model, which verifies the reasonableness and effectiveness of the method.

**Keywords:** Knowledge graph completion; Semantic information extraction; Word vector dimensionality reduction; Attention mechanism; Soft constraint rule;

## 1 Introduction[①]

With the advent of the era of big data and the continuous development of deep learning technology, the connection between data has become more and more of a research hotspot. The connotation of big data is no longer limited to data, but also includes knowledge and the composite of both. Therefore, it is no longer practical to model the data in today's world using traditional approaches. Knowledge graphs, a rapidly developing branch in the field of natural language processing, can construct symbolic knowledge into triads to solve many problems in the real world. Currently, knowledge graphs have played an important role in the fields of knowledge Q&A, intelligent search, recommender systems, and machine translation. Typical knowledge graphs represented by encyclopedic datasets include FreeBase,YAGO,WikiData,DBPetia,Nell,Probase, and Google's KnowledgeVault. Although existing knowledge graphs already contain thousands of entity relationships and triples, most existing knowledge graphs are sparse due to their inherent incompleteness, which gives rise to a new task: Knowledge Graph Complementation (KGC). The aim is to predict the missing triads in the knowledge graph.

In recent years, many solutions have been proposed by domestic and foreign experts and scholars for the task of knowledge graph complementation. The TransE model proposed by (Bordes, A et al., 2013) models the head and tail entities as vectors by representing the relations as translations in the embedding space, and optimizes the loss by "head entity + relation = tail entity". The model is simple, easy to train, and effective in a large number of datasets. However, its drawback is also obvious that in a one-to-many or many-to-many knowledge graph complementation scenario, there is a significant prediction bias due to the existence of multiple entities competing for a single point in embedding space. (Wang, Z et al., 2014) proposed the TransH model based on the TransE model. The idea of this model is to abstract the relations in the triad into a vector plane (Hyperplane), map the head node or tail node to this hyperplane each time, and then calculate the difference between the head and tail nodes by translating the vectors on the hyperplane. The main purpose of this is to compensate for the resulting bias of the TransE model in dealing with one-to-many or many-to-many relationships by introducing hyperplanes instead of the original relationship vectors so that the vector representations of the same node in different relationship hyperplanes are not identical. (Moon, C et al., 2017) proposed the TransR model, an approach that considers an entity as a composite of multiple attributes, different relations focus on different attributes of the entity, and different relations have different semantic spaces. Entities and relations are modeled in two different spaces, i.e., entity space and multiple relation spaces (relation-specific entity space), and transformations are performed in the corresponding relation spaces. Although the introduction of different semantic spaces enhances the effect of knowledge graph complementation, the parameters of its model increase dramatically, and the computational complexity is greatly improved. (Ji, G et al., 2015) proposed the TransD model based on the TransR model, which sets up two projection matrices, considers the diversity between entities and relationships, projects the head entity and tail entity into the relationship space, respectively, and uses two projection vectors to construct the projection matrix, solving the problem of too many parameters of the TransR model. (Xiao,H et al., 2015) proposed the TransA model to solve the problem of oversimplifying the loss metric and not having enough ability to measure and represent the diversity and complexity of entities and relationships in the knowledge base that exists in the translation-based knowledge representation approach.

Considering the use of semantic information in triads, (Yao, L et al., 2019) worked out a new knowledge graph complementation approach KG-BERT model, using the pre-trained language model BERT proposed by (Devlin, J et al., 2018) to complement the knowledge graph, treating the triads in the knowledge graph as text sequences, and proposing the new framework KG-BERT. The entity description and relationship description of the triad are used as inputs to calculate the triad scoring function using the KG-BERT model. The proposal of the KG-BERT model has led to new ideas in the study of knowledge graph complementation. However, since the BERT model requires replacing the head or tail entities with almost all entities for link prediction, it is very costly. Therefore a new solution StAR model was proposed by (Wang, B et al., 2021) using deterministic classifiers and spatial metrics for representation and structure learning. It avoids combinatorial explosion by reusing embeddings of graphical elements, thus reducing overhead, and enhances structured knowledge by exploring spatial features, obtaining significant results in experiments on knowledge graph complementation correlation. (Zhang, M et al., 2020) proposed a bidirectional encoder representation (BERT) model based on entity type information, which treats knowledge graph complementation as a classification task, where an entity, relationship, and entity type information is set into a text sequence, and Chinese characters are used as a token unit, and the model is pre-trained using a small amount of tokenized data, and then the pre-trained using a large amount of untagged data The model is then fine-tuned using a large amount of untagged data. The semantic information of the enriched entities was obtained well in experiments related to the knowledge graph complementation.

From the above-related research results, it can be concluded that the structural information and semantic information of the triad can play different roles in the process of knowledge graph complementation. At the structure level, the influence of triadic structural information on knowledge graph complementation is achieved by adding different constraints such as distance constraints, modal length constraints, orthogonality constraints, and other constraint rules through text encoding, graph embedding paradigm and so on. At the semantic level, word vectors with semantic information are obtained by introducing specific text encoding patterns and inputting triadic information into the pre-trained language model, and the knowledge graph is complemented by the generated word vectors. The structure-level complementation method only considers the triad structure information, which leads to the text semantics not being fully utilized and ignores the influence of semantic information on the knowledge graph complementation. Although the semantic level complementation method can vectorize the semantic information, it is inefficient in the knowledge graph complementation process due to the huge storage and computation overhead.

To address the above problems, this paper proposes a Knowledge Graph Completion method fused with Adaptive Enhanced Semantic Information (AESI-KGC). The main contributions of the method are reflected in the following points: (1) A way of extracting semantic features of triads is constructed, and triadic word vector encoding with semantic information is generated by fine-tuning the pre-trained language model. (2)Facing the high-latitude semantic information word vector, a dimensionality reduction method is constructed, which not only eliminates redundant features but also improves the efficiency of the model during training. (3) Construct the inter-triadic attention calculation formula by combining the generated semantic vector with the structural information vector, calculate the attention between relations and entities in positive and negative triads, and optimize the calculation results by the constructed semantic contrast loss function, adding them to the TransH model as constraints, and finally form the semantic information conditional constraints to realize the fusion and adaptive enhancement. (4) Comparison experiments are designed to verify that the proposed method in this paper has improved its evaluation metrics MR and Hits@10 in both knowledge graph datasets FB15K and WIN18 compared with the original translation model.

This paper is structured as follows: The Knowledge Graph Completion method fused with Adaptive Enhanced Semantic Information (AESI-KGC) is presented in Section 2. Section 3 presents the experiments and evaluation, including the experimental dataset, experimental evaluation metrics, and experimental results analysis. Section 4 concludes the paper, summarizes the whole paper, and provides an outlook.

## 2 Knowledge Graph Completion method fused with Adaptive Enhanced Semantic Information (AESI-KGC)

This paper proposes a Knowledge Graph Completion method fused with Adaptive Enhanced Semantic Information (AESI-KGC). The method consists of three parts, which are the extraction of semantic information from triads, dimensionality reduction of high-latitude word vectors, and construction of semantic information constraints. The overall method flow is shown in Figure 1. A positive sampling triple in the figure is represented as $(h, r, t)$, Where "$h$"represents the head entity, "$r$"represents the tail entity, and "$t$"represents the relationship connecting the two entities. Negative sampling triples are represented as $(h^`, r^`, t^`)$. The triple table with semantic information is $(h^*, r^*, t^*)$ and $(h^{`*}, r^{`*}, t^{`*})$. Dashed arrows indicate concatenation operations. Dashed arrows indicate concatenation operations. Text Embedding represents the semantic information vector generated by BERT, Parameter Embedding represents the initial parameter embedding vector to be learned,the P score represents the positive triplet semantic attention score,and the N score represents the negative triplet semantic attention score. The "Loss" is a loss function for calculating the semantic contrast information of positive and negative triples.

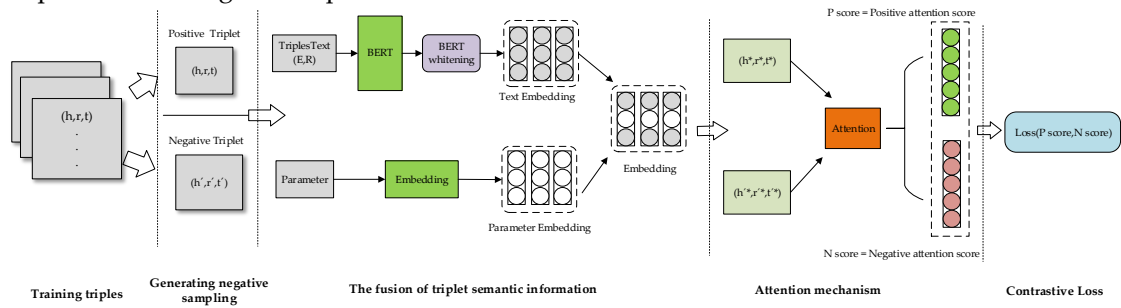

**Figure 1.** AESI-KGC Method Flow.

In order to obtain the semantic information representation of triplet words, the method inputs the entities and relations in the triplet into the pre-trained language model BERT for finetuning, and generates the semantic information word vector representation. Since high-latitude word vectors can impose a large burden on subsequent model training, by using the BERT-whitening dimensionality reduction method proposed by (Su, J et al., 2021) the high-dimensional vectors are dimensionally reduced using computational principles such as covariance, orthogonal matrix, etc., to eliminate redundant features and form a low-latitude semantic information word vector representation.The obtained semantic information is fused with the structural parameter information to form a vector representation of the triad, and by calculating the attention performance scores of relations to entities in the positive and negative triads, the output vector is input to the semantic information loss function for optimization to form semantic constraints added to the trnasH model, thus compensating for the lack of semantic information in the original model and making the knowledge map complementary effect improved.

*2.1 Translation-based distance model TransH*

The proposed approach in this paper is formed by adding semantic information constraints of triadic attention scores to the TransH model. The specific idea of this model is to abstract the relationship in the triplet into a hyperplane in a vector space and map the head node or tail node to this hyperplane every time. Then calculate the difference between the head and tail nodes through the translation vector on the hyperplane. As shown in Figure 2 : " $h$ "and" $t$ "vector represents the head node and tail node respectively. The relational hyperplane is represented by the normal vector of the plane and the translation vector on the plane.

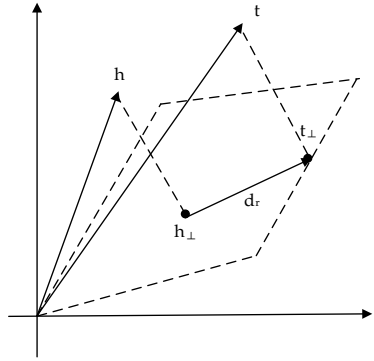

**Figure 2.** TransH model.

For a triple, we first need to map " $h$ " and " $t$ " to the hyperplane to get the mapping vectors " $h_\perp$ " and t" $t_\perp$ ". Specific formula (1)(2) indicates:

$$h_\perp = h - w_r^T h w_r \tag{1}$$

$$t_\perp = t - w_r^T t w_r \tag{2}$$

$w_r^T h = |w||h|\cos\theta$ represents the length of the projection of" $h$ "in the" $w_r$ "direction , multiplied by" $w_r$ ",that is, after the projection of " $h$ "on " $w_r$ "is obtained, the difference of the triplet can be obtained according to formula (3):

$$f_r(h,t) = ||(h - w_r^T h w_r) + d_r - (t - w_r^T t w_r)||_2^2 \tag{3}$$

The expectation of formula (3) is that if the triple relation is correct, the result of $f_r(h,t)$ will be smaller, and if not, the result will be larger. In order to meet the expectations proposed above, the model introduces a margin-base ranking function as a loss function to train the model, and the loss function is shown in (4):

$$L = \sum_{(h,r,t)\in\Delta} \sum_{(h',r',t')\in\Delta'_{(h,r,t)}} [f_r(h,t) + \gamma - f_{r'}(h',t')]_+ \tag{4}$$

Where $[x]_+$ is regarded as max(0, x), $\Delta$ represents the set of correct triples, and $\Delta'$ represents the set of negative triples. $\gamma$ is the margin value, which is used to distinguish positive and negative examples. The loss is trained by Mini-SGD, making $f_r(h,t)$ as small as possible and $f_{r'}(h',t')$ as large as possible when the training is out of date. In the process of minimizing the loss function, the model also proposes three soft constraint principles:

$$\forall e \in E, ||e||_2 \le 1 \qquad (5)$$

$$\forall r \in R, |w_r^T d_r| / ||d_r||_2 \in \varepsilon \qquad (6)$$

$$\forall r \in R, ||w_r||_2 = 1 \qquad (7)$$

Formula (5) is to ensure that the embeddings of all entities are normalized. Formula (6) is used to ensure that $w_r$ and $d_r$ are orthogonal and perpendicular to the $d_r$ super plane; Formula (7) ensures that the modulus of the normal vector is 1. In order to satisfy these three constraints, the original formula (4) needs to be changed, and formulas (5) and (6) need to be added. The changed loss function is shown in Formula (8):

$$L = \sum_{(h,r,t) \in \Delta} \sum_{(h',r',t') \in \Delta'_{(h,r,t)}} [f_r(h,t) + \gamma - f_{r'}(h',t')]_+ + C\{\sum_{e \in E}[||e||_2^2 - 1]_+ + \sum_{r \in R}[\frac{(w_r^T d_r)^2}{||d_r||_2^2} - \varepsilon^2]_+\} \qquad (8)$$

*2.2 Extraction of semantic information of triples*

Fusion of semantic information we use BERT to encode the semantic information of triples. The specific method is shown in Figure 3.

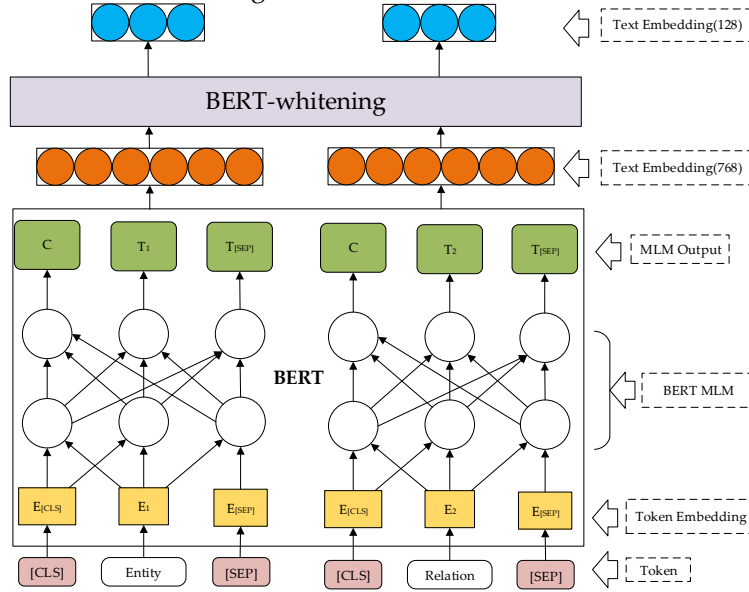

**Figure 3.** Triple semantic information extraction.

The input of BERT is the text information of Entity and Relation in the triple. We use string stitching to stitch "[CLS] + text information + [SEP]" into a string. The string is passed through a word dictionary to generate Tokens and the corresponding Token Embedding for each Token, which is constructed using the WordPiece algorithm. After that, the BERT masked language model (MLM) is used to generate a deep bi-directional language representation (MLM Output). In the figure, C is the output of the classification Token ([CLS]) corresponding to the last Transformer, and Ti represents the output of the other Tokens corresponding to the last Transformer. The word vectors of entities and relations are averaged over the penultimate hidden layer of each Token to generate a 768-length vector (Text Embedding(768)). This gives the semantic information vector generated by BERT.

2.2.1 Construction of word dictionary

The word dictionary is constructed using the WordPiece algorithm. The specific process of the algorithm is shown in Table 1:

Table 1. WordPiece Algorithm Workflow.

| Algorithm1 :WordPiece Algorithm |
| --- |
| 1: Prepare a large enough training corpus. |
| 2: Determine the desired subword word list size. |
| 3: Split the words into character sequences. |
| 4: Train the language model based on the data in step 3. |
| 5: Select the unit that maximizes the probability of training data after adding the language model as a new unit from all possible subword units. |
| 6: Repeat step 5 until the subword word list size or probability increment set in step 2 is reached below a certain threshold. |

Assuming that sentence $S = (t_1, t_2, \cdots t_n)$ consists of n subwords, $t_i$ represents subwords, and assuming that each subword exists independently, the language model likelihood value of the sentence S is equivalent to the product of the probabilities of all subwords, as shown in formula (9) shown:

$$\log P(S) = \sum_{i=1}^{n} \log P(t_i) \tag{9}$$

Assuming that the two subwords "x" and "y" in adjacent positions are merged, the resulting subword is denoted as "z", and the change in the likelihood value of sentence S can be represented by formula (10):

$$\log P(t_z) - (\log P(t_x) + \log P(t_y)) = \log(\frac{P(t_z)}{P(t_x)p(t_y)}) \tag{10}$$

From the above formulas (9) and (10), it can be concluded that the variation of the likelihood value is the mutual information between two words, and each time two subwords are selected for merging, they have the largest mutual information value, i.e., the two subwords are strongly related on the sublanguage model, and they often occur simultaneously in an adjacent manner in anticipation.

*2.3 Dimensionality reduction of high-dimensional word vectors*

In the above method, we obtained the semantic word vector generated by BERT. However, many studies have shown that the semantic vector obtained using BERT is not effective in similarity calculation, and the 768-dimensional word vector has great storage challenges for subsequent semantic attention calculation and parameter alignment. Considering the above problems, referring to the BERT-whitening model proposed by (Su, J et al., 2021), simple vector whitening is used to improve the word vector quality. It can not only improve the effect of semantic similarity calculation but also reduce the dimension of the semantic vector and improve the computational efficiency of semantic attention between subsequent triples. Referring to the calculation rules of the BERT-whitening model, we define a dimension transformation method, as shown in formula (11):

$$\tilde{x}_i = (x_i - \mu)W \tag{11}$$

Let the set of (row) vectors be $\{x_i\}_{i=1}^{N}$, and perform the transformation shown in formula (11) so that the mean of $\{x_i\}_{i=1}^{N}$ is 0 and the covariance is the identity matrix. There are two parameters $\mu$ and $W$ in the formula. Taking $\mu = \frac{1}{N}\sum_{i=1}^{N} x_i$ can make the mean of $\{x_i\}_{i=1}^{N}$ 0. The calculation method of parameter $W$ is: Denote the covariance matrix of the original data as $\Sigma$, and the covariance of the transformed data as $\tilde{\Sigma}$, so that the transformed data $U$ is an orthogonal matrix. $A$ is a bidiagonal matrix, and the diagonal matrix elements are all positive. According to the matrix transformation algorithm, the parameter $W = U\sqrt{A^{-1}}$ can be obtained.

After completing all the settings of the parameters of Formula (11), the semantic vector can be reduced in dimension. The specific process is shown in Table 2:

**Table 2.** Vector dimensionality reduction Workflow.

| Algorithm2 :Whitening-k Workflow |
| --- |
| **Input**: Existing embeddings $\{x_i\}_{i=1}^{N}$ and reserved dimensionality $k$ |
| 1: compute $\mu$ and $\Sigma$ of $\{x_i\}_{i=1}^{N}$ |
| 2: compute $U, \Lambda, U^T = SVD(\Sigma)$ |
| 3: compute $W = (U\sqrt{\Lambda^{-1}})[:,:k]$ |
| 4: **for** $i = 1, 2, \cdots, N$ **do** |
| 5: $\tilde{x}_i = (x_i - u)W$ |
| 6: **end for** |
| **Output:** Transformed embeddings $\{\tilde{x}_i\}_{i=1}^{N}$ |

Let k=128, the input is the triple semantic vector generated by BERT, and through the above algorithm, the final semantic vector representation is obtained. As shown in Figure 3, the 768-dimensional word vector (Text Embedding(768)) generated by BERT is reduced to 128-dimensional (Text Embedding(128)). This not only eliminates redundant features but also achieves the effect of speeding up and improving efficiency.

*2.4 Construction of Semantic Information Constraints*

The construction of semantic information constraints adopts the attention mechanism proposed by (Vaswani, A et al., 2017). By fusing the semantic information of the triplet with the parameter vector of the original model, it is input into the Attention structure. Computing attention scores for semantic information of triples. The specific operation is shown in Figure 4.

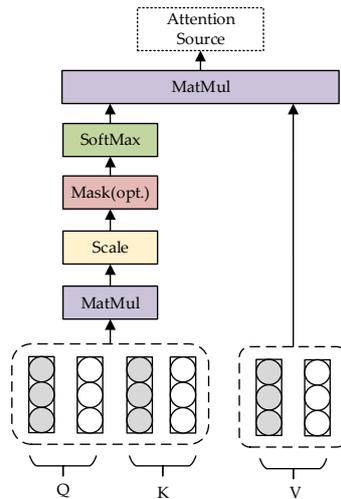

**Figure 4.** Attention scores for fused semantics

In the figure, the gray circle is the semantic information vector, and the white circle is the parameter vector. The semantic information representation about the parameter is obtained by adding the two together. Q is the fusion of the head entity vector and semantic information vector, K is the fusion of semantic information of the tail entity vector, and V is the fusion of the relation vector and semantic information. Q and K are multiplied to obtain the relationship weight between each word vector, which is multiplied by the scaling factor $\frac{1}{\sqrt{d_k}}$. After normalization by softmax, the attention output is finally obtained by multiplying with V. The specific calculation method is shown in formula (12):

$$Attention(Q,K,V)=softmax(\frac{QK^T}{\sqrt{d_k}})V \qquad (12)$$

The positive and negative triples get the attention scores on semantic information after the above attention operations. After obtaining the scores of the two, we adopt a new semantic information contrast loss function, as shown in formula (13) (14) Show:

$$l_{contr}^{x_i x_j}(W) = \frac{e^{sim(SiTcontr(x_i),SiTcontr(x_j))/\tau}}{\sum_{k=1,k\neq i}^{2N} e^{sim(SiTcontr(x_i),SiTcontr(x_k))/\tau}} \qquad (13)$$

$$L = -\frac{1}{N}\sum_{j=1}^{N} \log l_{contr}^{x_j x_{-j}}(W) \qquad (14)$$

$x_i$ and $x_j$ represent the semantic score of the positive and negative triples of the input respectively, $sim(\cdot,\cdot)$ represents the cosine similarity, that is, the dot product after normalization, and $\tau$ is the over-temperature parameter, which is used to adjust the degree of attention to difficult samples, the smaller the overtemperature parameter focuses more on separating this sample from the most similar other samples. $x_j$, $x_{-j}$ represent input vectors from two different data augmentations under the same semantics. $S_iTcontr(\cdot)$ denotes the semantic information representation obtained from the comparison.

$$L = \sum_{(h,r,t)\in\Delta}\sum_{(h',r',t')\in\Delta'_{(h,r,t)}} [f_r(h,t)+\gamma - f_{r'}(h',t')]_+$$
$$+C\{\sum_{e\in E}[||e||_2^2 - 1]_+ + \sum_{r\in R}[\frac{(w_r^T d_r)^2}{||d_r||_2^2} - \varepsilon^2]_+ + -\frac{1}{N}\sum_{j=1}^{N}\log l_{contr}^{x_j x_{-j}}(W)\} \qquad (15)$$

Through this method, the attention constraint of semantic information is added un-der the original soft constraint rules, so that different triples can obtain their semantic in-formation. After adding to the original model, the semantic information of triples can be adapted for enhancement.

## 3 Experiment and Evaluation

*3.1 Experimental Dataset*

To evaluate the proposed method, two datasets are used for experiments, namely WIN18 and FB15k. Table 3 provides statistics on the datasets used in the experiments.

Table 3. Statistics of WN18 dataset and FB15K dataset.

| Dataset | Entities | Relations | #Edges | | | |
|---|---|---|---|---|---|---|
| | | | Training | Validation | Test | Total |
| WN18 | 40,943 | 18 | 141,442 | 5000 | 5000 | 151,442 |
| FB15K | 14,951 | 1,345 | 483,142 | 50,000 | 59,071 | 592,213 |

WN18 and FB15K have relatively rich triples, WN18 has a total of 40,943 entities and 18 relationships. FB15k has a total of 14,951 entities with 1,345 relationships. WN18 and FB15K suffer from the inverse relation problem in relation prediction tasks, so state-of-the-art results can be obtained using a simple inversion rule-based model.

*3.2 Experimental Evaluation Metrics*

In the relation prediction task, the goal is to predict the triplet (h, r, t) for which h or t is missing, predict given (r, t) to predict "h" or given (h, r) to predict "t". The specific experimental process is as follows: Generates a set of (N−1) triples, corrupting each entity's triples by replacing each head entity "h" with every other triple head entity "h". Then assign a score to each such triple. These scores are then sorted in ascending order and the ranking of the correct triples is obtained. All models were evaluated in a filtering setting, that is, during sorting, with corrupt triples already present in one of the train, validation or test sets removed. Repeat the whole process by replacing the tail entity and report the average metric. During the experiment, the mean rank (MR) and the proportion of correct entities (Hits@10) among the top N are evaluated. Hits@10 refers to the average proportion of triples that rank less than 10 in link prediction. The specific calculation method is shown in formulas (16) and (17):

$$MR = \frac{1}{|S|}\sum_{i=1}^{|S|} rank_i = \frac{1}{|S|}(rank_1 + rank_2 + \ldots + rank_{|S|}) \tag{16}$$

$$HITS@n = \frac{1}{|S|}\sum_{i=1}^{|S|} \prod(rank_i \leq n) \tag{17}$$

Where S is the triplet set, |S| is the number of triplet sets, $rank_i$ is the link prediction ranking of the i-th triplet, n=10, and $\prod$ represents the indicator function (if the condition is true, the function value is 1, otherwise 0).

*3.3 Experimental Results Analysis*

We experimentally investigate and evaluate two related tasks on the data, link prediction experiment and ablation experiments with model hyperparameter settings.

3.3.1. Link Prediction Experiment

In the knowledge graph, tail entities account for the vast majority of entities, and the same head entity and relationship may connect thousands of tail entities. Faced with such a huge amount of data, when link prediction is performed, if there are damaged triples in the knowledge graph, the result of link prediction will be biased. To eliminate this factor, corrupt triples present in the training, valid and test sets are removed before each test triple is obtained. The removal of the damage triplet setting is called "Filt", and the one that is not removed is set to "Raw". In both cases, lower MR values are better, while higher Hits@10 values are better.

Table 4 shows the experimental results of the WN18 and FB15K test sets. We experimentally compared the four models respectively. The Hits@10 value is expressed as a percentage. The best score is in bold, the next best score is underlined.

**Table 4** Comparison of experimental results of link prediction between WN18 and FB15K datasets

| Datsset | WN18 | | | | FB15K | | | |
|---|---|---|---|---|---|---|---|---|
| Metirc | MR | | HITS@10 | | MR | | HITS@10 | |
| | Raw | Filt | Raw | Filt | Raw | Filt | Raw | Filt |
| LFM | 469 | 456 | 71.4 | 81.6 | 283 | 164 | 26.0 | 33.1 |
| TransE | **263** | **251** | <u>75.4</u> | 89.2 | 243 | 125 | 34.9 | 47.1 |
| TransH(unif) | 318 | 303 | <u>75.4</u> | <u>86.7</u> | 211 | <u>84</u> | 42.5 | <u>58.5</u> |
| TransH(bern) | 400.8 | 388 | 73.0 | 82.3 | <u>212</u> | 87 | <u>45.7</u> | 46.4 |
| AESI-KGC | <u>302.2</u> | <u>300.9</u> | **77.3** | **87.9** | **201** | **82** | **48.3** | **59.8** |

According to Table 4, the transE model performed the best MR test score in the WN18 dataset with a small number of relationships, followed by the AESI-KGC method. However, compared to the TransH model, the AESI-KGC method was numerically improved and obtained suboptimal results. In the test experiments with HITS@10, the model with the AESI-KGC method added improved the experimental results over the original model TransH by 2.3% in the value of raw and 1.2% in the value of Filt. In the FB15K dataset, where the number of relationships is much higher, TransH's model performs poorly and is inferior to the AESI-KGC method. Compared with the TransH model, the model with the addition of the AESI-KGC method shows a certain degree of improvement in the MR as well as the HITS@10 experimental values. In MR's experiments the Raw condition boosted by 10 and the Filt condition boosted by 5. In the HITS@10 experiment, the boost was 2.6% in Raw condition and 1.3% in Filt condition.

3.3.2. Ablation experiment

In the design of the ablation experiments, we mainly compare the loss function curves of this model under different conditions and select the best experimental hyperparameters from the comparison results by comparing the performance of the loss function for the parameters of the learning rate, epoch size, and embedding dim. The following describes the ablation comparison experiments on the WN18 dataset and FB15K dataset, respectively.

The four experimental results of the WN18 data set are shown in Figure 5:

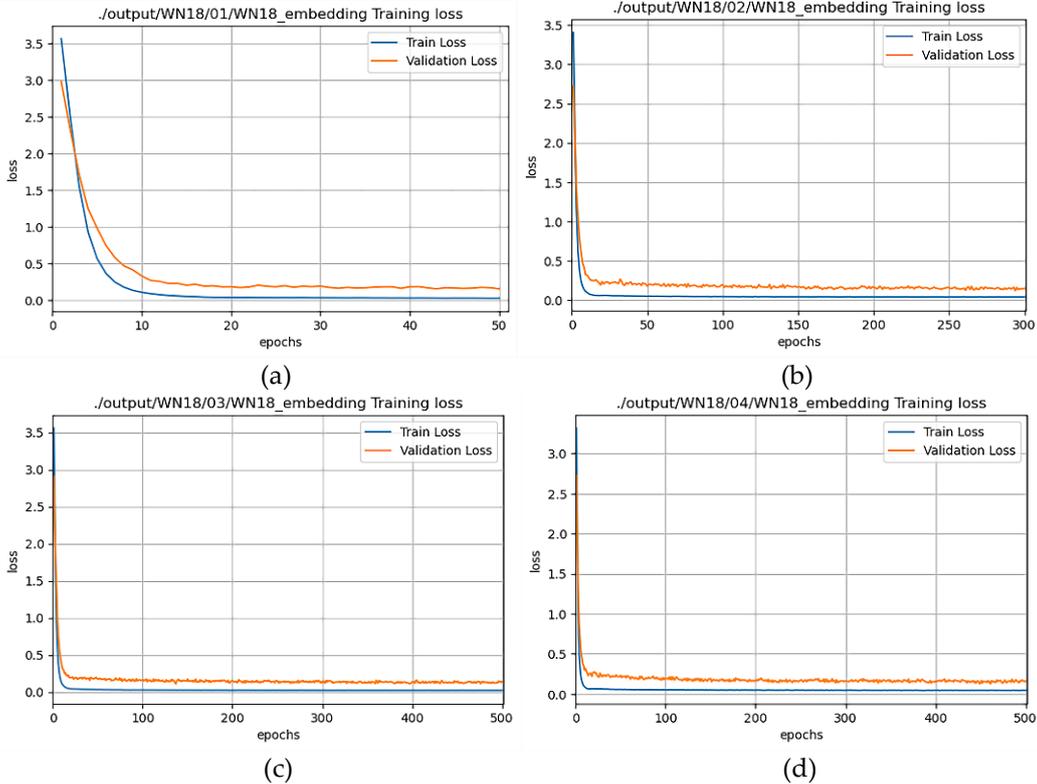

**Figure 5.** WN18 Dataset Ablation Experiment

The parameters in Figure 5-(a) are lr=0.001, margin=4.0, norm=1, C=0.001, embedding dim=256, epochs=50. At this point, it can be observed that the Train Loss decreases smoothly between 0-20, and there are slight fluctuations in the range of 20-50. Considering the number of epochs used in the experiment, the experiment in Figure 5-(b) keeps other conditions unchanged and increases the number of epochs to 300. It is observed that with the increase of epochs, the decrease of loss does not increase significantly. It can be seen from this that the WIN18 data set has a large change in the first 20 epochs under the above parameters, and the subsequent loss tends to be stable with the increase of epoch.

The comparison in Figure 5-(c) and Figure 5-(d) shows the vector dimension size used in this experiment. Since the native BERT outputs a 768-dimensional vector size, two vector size plans are adopted when performing dimensionality reduction on the output vectors, one for 256 dimensions as shown in Figure 5-(c) and one for 128 dimensions as shown in Figure 5-(d). It can be seen from the figure that when performing loss gradient descent, the Tran Loss of 256 dimensions starts from above 3.5, while 128 dimensions start from below 3.5. In terms of the time dimension, the 128-dimensional vector takes less time to train the model and achieves the same training effect as the 256-dimensional vector, so we use the 128-dimensional vector standard to reduce the semantic information derived from the pre-trained language model BERT under both training effects. Finally, the experimental Train Loss is stabilized at about 0.01 and Validation Loss is stabilized at about 0.18.

The four experimental results of the FB15K data set are shown in Figure 6:

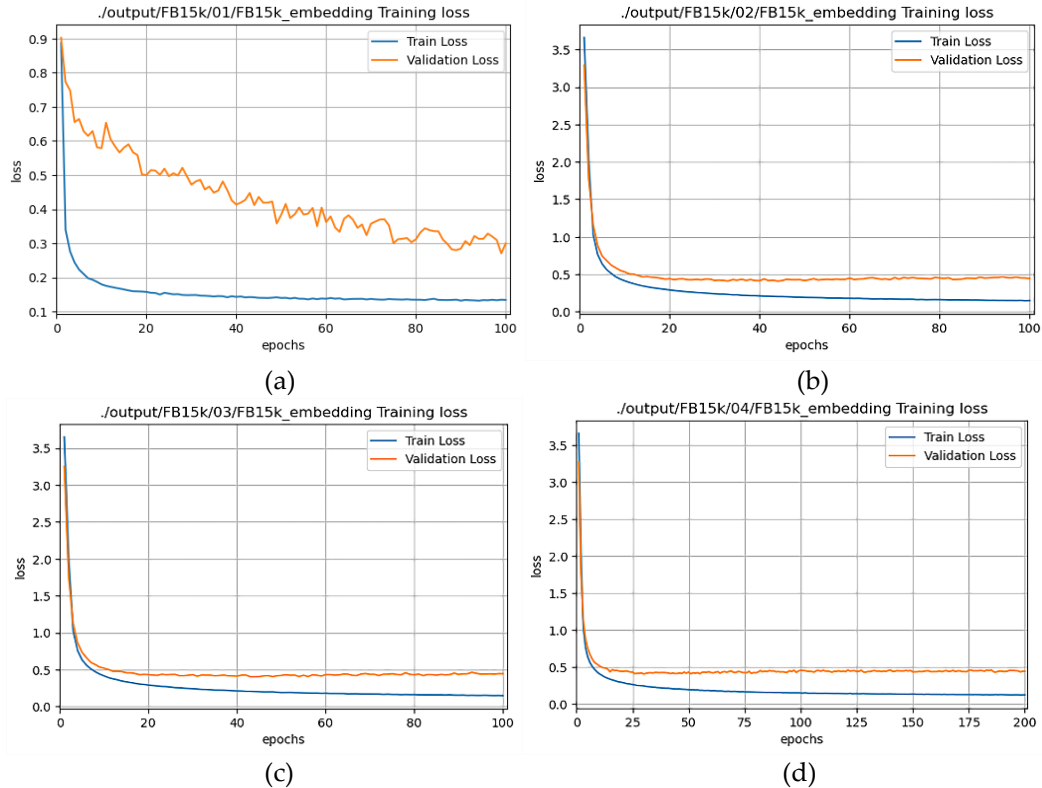

**Figure 6.** FB15K Dataset Ablation Experiment

The parameters of the experiment in Figure 6-(a) are embedding dim=128, lr=0.0015, margin=3.0, norm=1, C=0.001, epochs =100. From Figure 6-(a), it can be seen that during the experiment, Train Loss can keep decreasing smoothly, Validation Loss has obvious ups and downs and is unstable. After debugging, the parameters were changed to embedding dim=128, lr=0.001, margin=4.0, norm=1, C=0.0015, epochs =100. The experimental results are shown in Figure 6-(b), and it can be seen that the Validation Loss tends to be stable in the process of decreasing.

In the next experiment, we adjust C=0.002 in the above experimental parameters to increase the effect of soft constraint on the experimental effect. The trend of its loss function is shown in Figure 6-(c). In the last experiment, we add the calculation of semantic cosine similarity to the comparison function in the experiment while changing the number of epochs, and the effect is shown in Figure 6-(d). Since the FB15K dataset has more relations than the WN18 dataset, more triads are formed and the training time is greatly increased. The final experimental Train Loss is stabilized at about 0.16 and the Validation Loss is stabilized at about 0.5.

## 4 Conclusions

In this paper, we propose a Knowledge Graph Completion method fused with Adaptive Enhanced Semantic Information(AESI-KGC). The ternary word vectors with semantic information are generated by the pre-trained language model BERT, a semantic vector dimensionality reduction method is proposed for the problem of high feature dimensionality, and a new semantic soft constraint is constructed by increasing the contrast loss function of attention scores of positive and negative ternary semantic information. The method is added to the TransH model for experiments. The experimental results show that the original model is further improved by adding the constraint of semantic information in the triad, making full use of the semantic information in the triad, and effectively improving the success rate of the TransH model's complementation. However, with the increase of relations in KG, the semantic information may be conflicting, which will lead to the failure of correct complementation of the knowledge graph successfully. Therefore, in the face of multi-relational triples and similar semantic triples, it will be the next research direction to construct new strategies to cope with similar semantic triples, and then improve the efficiency of knowledge graph complementation.

**Author Contributions:** Conceptualization, G.Z. and W.J.; methodology, Z.Y. and W.J.; software, Y.Y. and X.Z.; validation, Y.Y. ,X.Z. and C.S; formal analysis, C.S.;data curation, Z.Y. and X.Z .; writing—original draft preparation, Z.Y.; writ-ing—review and editing, W.J. and Z,Y; visualization, C.S.;supervision, G.Z. and W.J.; project ad-ministration, Z.Y. and Y.Y.; funding acquisition, W.J. and G.Z. All authors have read and agreed to the published version of the manuscript.

**Funding:** This research was funded by the National Natural Science Foundation of China, grant number 31971015 and funded by Natural Science Foundation of Heilongjiang Province in 2021 under, grant no LH2021F037.

## ORCID

| | |
|---|---|
| Weidong Ji | https://orcid.org/0000-0002-0850-6538 |
| Zengxiang Yin | https://orcid.org/0000-0003-0749-9740 |
| Guohui Zhou | https://orcid.org/0000-0002-5162-9058 |
| Yuqi Yue | https://orcid.org/0000-0003-0522-2617 |
| Xinru Zhang | https://orcid.org/0000-0002-4172-0691 |
| Chenghong Sun | https://orcid.org/0000-0002-5930-8128 |

## REFERENCES

Atito, S., Awais, M., & Kittler, J. (2021). Sit: Self-supervised vision transformer. arXiv preprint arXiv:2104.03602.
doi: 10.48550/arXiv.2104.03602

Bizer, C., Lehmann, J., Kobilarov, G., Auer, S., Becker, C., Cyganiak, R., & Hellmann, S. (2009). Dbpedia-a crystallization point for the web of data. Journal of web semantics, 7(3), 154-165.
doi: 10.1007/978-3-540-76298-0_52

Bordes, A., Usunier, N., Garcia-Duran, A., Weston, J., & Yakhnenko, O. (2013). Irreflexive and hierarchical relations as translations. arXiv preprint arXiv:1304.7158.
doi: 10.48550/arXiv.1304.7158

Bollacker, K. . (2008). Freebase : a collaboratively created graph database for structuring human knowledge. Proc. SIGMOD' 08. 1247-1250.
doi: 10.1145/1376616.1376746

Devlin, J., Chang, M. W., Lee, K., & Toutanova, K. (2018). Bert: Pre-training of deep bidirectional transformers for language understanding. arXiv preprint arXiv:1810.04805.
doi: 10.48550/arXiv.1810.04805

Dong, X., Gabrilovich, E., Heitz, G., Horn, W., Lao, N., & Murphy, K. (2014). omas Strohmann, Shaohua Sun, and Wei Zhang. 2014. Knowledge vault: a web-scale approach to probabilistic knowledge fusion. In Proc. of KDD (pp. 601-610).
doi: 10.1145/2623330.2623623

Isinkaye, F. O., Folajimi, Y. O., & Ojokoh, B. A. (2015). Recommendation systems: Principles, methods and evaluation. Egyptian informatics journal, 16(3), 261-273.
doi:10.1016/j.eij.2015.06.005

Ji, G., He, S., Xu, L., Liu, K., & Zhao, J. (2015, July). Knowledge graph embedding via dynamic mapping matrix. In Proceedings of the 53rd annual meeting of the association for computational linguistics and the 7th international joint conference on natural language processing (volume 1: Long papers) (pp. 687-696).
doi: 10.3115/v1/P15-1067

Longpre, S., Perisetla, K., Chen, A., Ramesh, N., DuBois, C., & Singh, S. (2021). Entity-based knowledge conflicts in question answering. *arXiv preprint arXiv:2109.05052*.
doi: 10.48550/arXiv.2109.05052

Miller, G. A. (1995). WordNet: a lexical database for English. Communications of the ACM, 38(11), 39-41.
doi: 10.1145/219717.219748

Moon, C., Jones, P., & Samatova, N. F. (2017, November). Learning entity type embeddings for knowledge graph completion. In Proceedings of the 2017 ACM on conference on information and knowledge management (pp. 2215-2218).
doi: 10.1145/3132847.3133095

Suchanek, F. M., Kasneci, G., & Weikum, A. G. (2008). Yago - a large ontology from wikipedia and wordnet. Web Semantics Science Services & Agents on the World Wide Web, 6(3), 203-217.
doi: 10.1016/j.websem.2008.06.001

Schuster, M., & Nakajima, K. (2012, March). Japanese and korean voice search. In 2012 IEEE international conference on acoustics, speech and signal processing (ICASSP) (pp. 5149-5152). IEEE.
doi: 10.1109/ICASSP.2012.6289079

Su, J., Cao, J., Liu, W., & Ou, Y. (2021). Whitening sentence representations for better semantics and faster retrieval. arXiv preprint arXiv:2103.15316.
doi: 10.48550/arXiv.2103.15316

Taylor, W. L. (1953). "Cloze procedure": A new tool for measuring readability. Journalism quarterly, 30(4), 415-433.
doi: 10.1177/107769905303000401

Tohidi, M., Khayat, N., & Telvari, A. (2022). The use of intelligent search algorithms in the cost optimization of road pavement thickness design. Ain Shams Engineering Journal, 13(3), 101596.
doi:10.1016/j.asej.2021.09.023

Undavia, J. N., Patel, A., & Patel, S. (2021). Security issues and challenges related to Big Data. *Research Anthology on Privatizing and Securing Data*, 1604-1620.
doi:10.4018/978-1-7998-8954-0.ch077

Vaswani, A., Shazeer, N., Parmar, N., Uszkoreit, J., Jones, L., Gomez, A. N., ... & Polosukhin, I. (2017). Attention is all you need. Advances in neural information processing systems, 30.
doi: 10.48550/arXiv.1706.03762

Vrandečić, D., & Krötzsch, M. (2014). Wikidata: a free collaborative knowledgebase. Communications of the ACM, 57(10), 78-85.
doi: doi.org/10.1145/2629489

Wang, B., Shen, T., Long, G., Zhou, T., Wang, Y., & Chang, Y. (2021, April). Structure-augmented text representation learning for efficient knowledge graph completion. In Proceedings of the Web Conference 2021 (pp. 1737-1748).
doi: 10.1145/3442381.3450043

Wang, H., Wu, H., He, Z., Huang, L., & Church, K. W. (2021). Progress in Machine Translation. Engineering.
doi:10.1016/j.eng.2021.03.023

Wang, Q., Mao, Z., Wang, B., & Guo, L. (2017). Knowledge graph embedding: A survey of approaches and applications. IEEE Transactions on Knowledge and Data Engineering, 29(12), 2724-2743.
doi:10.1109/TKDE.2017.2754499

Wang, Z., Zhang, J., Feng, J., & Chen, Z. (2014, June). Knowledge graph embedding by translating on hyperplanes. In Proceedings of the AAAI conference on artificial intelligence (Vol. 28, No. 1).
doi: 10.1609/aaai.v28i1.8870

Wu, W., Li, H., Wang, H., & Zhu, K. Q. (2012, May). Probase: A probabilistic taxonomy for text understanding. In Proceedings of the 2012 ACM SIGMOD international conference on management of data (pp. 481-492).
doi: 10.1145/2213836.2213891

Xiao, H., Huang, M., Hao, Y., & Zhu, X. (2015). TransA: An adaptive approach for knowledge graph embedding. arXiv preprint arXiv:1509.05490.
doi: 10.48550/arXiv.1509.05490

Yao, L., Mao, C., & Luo, Y. (2019). KG-BERT: BERT for knowledge graph completion. arXiv preprint arXiv:1909.03193.
doi: 10.48550/arXiv.1909.03193

Zhang, M., Geng, G., Zeng, S., & Jia, H. (2020). Knowledge Graph Completion for the Chinese Text of Cultural Relics Based on Bidirectional Encoder Representations from Transformers with Entity-Type Information. Entropy, 22(10), 1168.
doi: 10.3390/E22101168

**AUTHOR BIOGRAPHIES**

**Weidong Ji** was born in Heilongjiang Province, China, in 1978. He received the M.S degree in Control Theory and Control Engineering from Harbin University of Science and Technology in 2004, and the Ph.D. degree in Mechanical Design and Theory from Northeast Forestry University in 2013. He is now a professor in the College of Computer Science and Information Engineering, Harbin Normal University. His research interests include Big data and swarm intelligence.

**Zengxiang Yin** was born in Henan Province, China in 1996. He received his bachelor's degree in engineering from Zhengzhou University of Light Industry in 2019 and his master's degree in electronic information from Harbin Normal University in 2021. His research interests include natural language processing and knowledge graph.

**Guohui Zhou** was born in Harbin in 1973, Ph.D, professor. His research interests include Machine vision and pattern recognition, knowledge mapping, educational evaluation, etc.

**Yuqi Yue** was born in 1998 in Heilongjiang Province, China. she received her B.S. degree in Computer Science and Technology from Harbin Normal University in 2020. She is now a master's student in the School of Computer Science and Information Engineering at Harbin Normal University. Her research

interests include population intelligence and knowledge tracing

**Xinru Zhang** was born in 1999 in Heilongjiang Province, China. She received her B.S. degree in Computer Science and Technology from Harbin Normal University in 2021. She is now a master's student in the School of Computer Science and Information Engineering at Harbin Normal University. Her research interests include artificial intelligence and multi-objective optimization.

**Chenghong Sun** was born in Shandong Province, China, in 1998. He received the Bachelor degree in Information engineering from Shandong University of Science and Technology in 2021. He is now a Master's degree student in the College of Computer Science and Information Engineering, Harbin Normal University. His research interests include Natural language processing and Knowledge graph.

**FOOTNOTE**

①: The current version of the manuscript is a revision of the content as well as the format of the previous preprint. Compared with the original preprint, its main novelties are shown as follows: (1) The biography of the author of this paper has been added. (2) The author's ORCID has been added. (3)And the format of literature citation in the introduction section has been modified. There is no change of author personnel in the current version.